\documentclass[letterpaper, 10 pt, conference]{ieeeconf}  

\IEEEoverridecommandlockouts                              

\usepackage{algorithm}
\usepackage{algpseudocode}
\usepackage{graphicx}
\usepackage[caption=false,font=footnotesize]{subfig}
\usepackage[hidelinks,colorlinks=true,linkcolor=blue,citecolor=blue]{hyperref}
\usepackage{amsmath}
\usepackage{amsfonts}
\usepackage{amssymb}
\usepackage{booktabs}
\usepackage{multirow}
\usepackage[capitalise]{cleveref}
\crefname{section}{Sec.}{Secs.}
\crefname{table}{Tab.}{Tabs.}
\usepackage{amsmath} 

\newcommand{\indexUAV}{m}
\newcommand{\typeUAV}{M}

\newcommand{\numUAVs}{N_{\typeUAV}}
\newcommand{\numGraphEdge}{\,|\,\EdgeSet\,|\,}
\newcommand{\numGraphNode}{\,|\,\NodeSet\,|\,}

\newcommand{\commRadius}{r_c}

\newcommand{\subregionsName}{GES}

\newcommand{\graphUAV}{G^{\indexUAV}}
\newcommand{\graphUAVext}{\graphUAV_{ext}}

\newcommand{\NodeSet}{V}
\newcommand{\EdgeSet}{E}

\newcommand{\graphNodeSet}{V^\indexUAV}
\newcommand{\graphEdgeSet}{E^\indexUAV}
\newcommand{\node}{v}
\newcommand{\edge}{e}

\newcommand{\vHType}{H}
\newcommand{\vFTypeAny}{F}
\newcommand{\vFTypeZero}{N}
\newcommand{\vFTypeOne}{F}

\newcommand{\nodeFrontier}{\node_{\vFTypeOne}}
\newcommand{\nodeFrontierZero}{\node_{\vFTypeZero}}
\newcommand{\nodeFrontierAdj}{\node_{\vFTypeZero}^{adj}}

\newcommand{\nodeHistorical}{\node_{\vHType}}
\newcommand{\nodeAgent}{\node_{\typeUAV}}

\newcommand{\edgeHH}{\edge_{\vHType\vHType}} 
\newcommand{\edgeHF}{\edge_{\vHType\vFTypeAny}} 
\newcommand{\edgeFF}{\edge_{\vFTypeAny\vFTypeAny}} 
\newcommand{\penaltyFF}{\rho_{FF}} 

\newcommand{\parentFset}{P_F} 
\newcommand{\parentHset}{P_H} 

\newcommand{\uavPose}{p^\indexUAV} 
\newcommand{\uavNode}{\nodeAgent^\indexUAV}

\newcommand{\matrixAdj}{M_{adj}}
\newcommand{\kmeansSteps}{T}

\title{\LARGE \bf
Connectivity-Aware Graph Extension for Decentralized Multi-Robot Exploration
}

\author{B\'eatrice Garcia Cegarra$^{\star \dagger}$, Elena Vanneaux$^{\dagger}$, Quentin Picard$^{\star}$ and David Filliat$^{\star}$%
\thanks{$^{\star}$P\^ole recherche, AMIAD, Palaiseau, France}%
\thanks{$^{\dagger}$UI2S, ENSTA, IP Paris, Palaiseau, France}%
\thanks{All authors are members of LARIAD, a joint AMIAD-ENSTA laboratory}
}

\begin{document}

\maketitle
\thispagestyle{empty}
\pagestyle{empty}

\begin{abstract} 

Exploring unknown environments with multiple UAVs requires coordination under intermittent communication, making decentralized operation a baseline assumption. We propose, within a decentralized framework, a novel exploration graph extension strategy based on frontier connectivity to extend exploration plans and maintain area partitioning among agents stable and robust to disconnections and changes in spatial layout. The proposed extension method is applied to two state-of-the-art area partitioning methods and evaluated in simulation. Experiments show improved performance over existing graph extension approaches with higher exploration efficiency under low communication rate.

\end{abstract}


\section{INTRODUCTION}

Exploring unknown environments is a core task in autonomous robotics, and extending it to multi-agent systems introduces the key challenge of balancing communication constraints with fast, coordinated behavior. Since UAV communication is limited in range, frequency, and reliability due to obstacles or interference, we treat intermittent connectivity as a baseline condition rather than an exception, and we aim at studying and mitigating its impact on exploration efficiency. We focus on distributed decision-making where each UAV acts locally based on partial observations enriched by occasional encounters, thus implementing indirect and asynchronous collaboration.

Multi-robot exploration (MRE) is a specification of the multi-robot task allocation problem (MRTA). Tasks are exploration areas that are allocated among agents and the main motivation of this work is to understand how structured partial information can be leveraged to ensure both efficient wide-ranging exploration and consistent task allocation across UAVs operating without reliable communication. The objective is to use this structure to align decisions across UAVs, reduce variability in task allocation, and guide exploration toward structurally meaningful and distant regions while limiting inefficient revisits.

To address this problem, we adopt a decentralized framework where each UAV makes decisions from a shared, partially observed topological graph to structure regions connectivity. The graph is further partitioned with Voronoi \cite{dong_fast_2024} or K-medoids \cite{goodwin_k-means_2022}. As our main contribution, we propose a novel exploration graph extension method based on frontier addition and connectivity preservation. This method aims to capture higher-level exploration plans and maintain a partition robust to disconnections or changes in spatial layout—i.e., avoiding abrupt changes in the resulting partition shape—and staying consistent in assigning subregions across drones and time without explicit communication over allocation stages. 

We adopt a mechanism similar to coverage nodes in \cite{dingBalancedCollaborativeExploration2025}, which use a UAV-centered sliding window to sample additional exploration targets beyond immediate reach. The key difference of our approach is its awareness of graph topology and extension from all frontiers. In contrast to sampling-based methods that generate candidate nodes within a local range-limited window around each UAV, we introduce nodes over the entire graph domain. These nodes are added deterministically based on topological parent-set intersection as an extension criterion. Finally, we evaluate both extensions across the two partitioning methods in two topologically distinct environments under limited communication, using a full-range communication setting as a reference.

This paper is organized as follows: first, the multi-agent exploration strategies with a focus on task allocation are reviewed in \cref{sec_stateart}. Then, the graph structure, communication scheme, and local exploration pipeline from graph partitioning to planning are detailed in \cref{sec_methods}. Finally, the exploration performances, such as connectivity and partition robustness across two environments using different partitioning methods and extension levels are evaluated in \cref{sec_exp_result}.

\begin{figure}[!t]
    \centering
    \includegraphics[width=\linewidth]{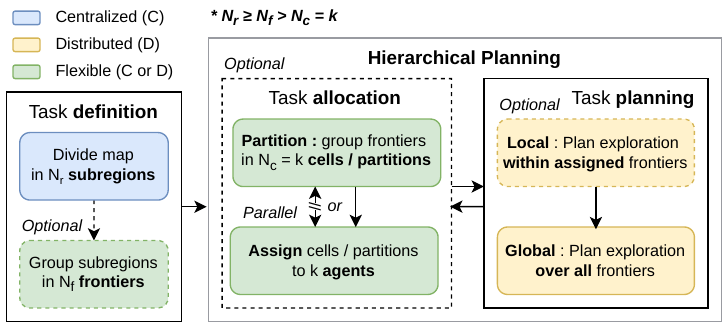}
    \caption{Iterative task-driven MRE workflow}
    \label{fig:frontier_samp}
\end{figure}

\section{RELATED WORK}
\label{sec_stateart}

\cref{fig:frontier_samp} illustrates the complete MRE pipeline, with each stage detailed in the following sections. In unknown environments, the exploration area may be either dynamically expanding or predefined and subsequently partitioned into a grid of sub-regions. MRE methods operate either on individual subregions or on aggregates of subregions formed via continuity \cite{yu_smmr-explore_2021}, \cite{yamauchiFrontierbasedExplorationUsing1998} similarity, proximity \cite{gaoMeetingMergingMissionMultirobotCoordinate2022} or density criteria \cite{ribeiroEfficient3DExploration2024}. In \cite{ribeiroEfficient3DExploration2024} and \cite{tang_novel_2020}, mean-shift and K-means enable scalable geometric and density-based clustering, while~\cite{gaoMeetingMergingMissionMultirobotCoordinate2022} uses spectral clustering for higher-dimensional grouping. Subregion cluster count varies with exploration progress and environment topology. However, clustering introduces an additional level of abstraction, enabling more informed MRTA decisions while reducing computational cost.

\subsection{Frontier-based exploration (FBE)}
\label{ssec_fbe}

The two most widely used approaches to MRE adapt single-agent exploration strategies based on frontier detection or sampling. Frontier-based approaches \cite{yamauchiFrontierbasedExplorationUsing1998} guide agents toward frontiers, i.e., subregions at the interface between explored and unexplored space, ensuring complete coverage with lower computational cost. Sampling-based approaches, in contrast, select next agent positions based on information gain evaluated over a random set of sampled points. Subsequent exploration methods are exclusively frontier-based.

Utility-based methods evaluate frontiers using a scalar scoring function, considering factors like distance, frontier size \cite{lupertoMultirobotRendezvousCommunicationrestricted2025}, turning cost \cite{gao_improved_2018}, density, feasibility \cite{ribeiroEfficient3DExploration2024}, and potential connectivity loss or recovery \cite{benavidesAutoAdaptiveMultiObjectiveStrategy2019}, \cite{benavides_multi-robot_2016}. Utility functions can be replaced by longer-term  sequential foresight methods that optimize trajectories but have high computational costs such as the Vehicle Routing Problem \cite{gaoMeetingMergingMissionMultirobotCoordinate2022} where frontiers are used as intermediate way-points. Potential field approaches use attractive forces from frontiers and repulsive forces from obstacles and other agents position and tasks to continuously guide exploration via gradient-based motion \cite{pongsirijindaMEFExploreCommunicationConstrainedMultiRobot2025}, \cite{zhangMRTopoMapMultiRobotExploration2022}. These methods are suitable for dynamic environments but less predictable due to local minima tackled by wavefront-based distance transforms \cite{yu_smmr-explore_2021}.

\subsection{Area partitioning for exploration}
\label{ssec_partition_methods}

Some works allocate cells or sets of frontiers to UAVs rather than individual frontiers, reducing computational cost and enabling longer-horizon planning. This improves performance over single-stage FBE algorithms detailed below by yielding more stable assignments and a more spatially distributed swarm. Within each cell or subregion, frontiers are explored sequentially during planning, following utility-based principles (e.g. current load, UAV velocity). Allocation denotes the stage comprising both space partition into clusters and their assignment to agents. 

Voronoi diagrams (VG) are a popular allocation approach due to low computational cost and suitability for distributed implementations. This method partitions the environment into dynamic cells, or equivalently assigns each frontier to an agent, typically the spatially closest relative to all others \cite{dong_fast_2024}, \cite{dingBalancedCollaborativeExploration2025}. Agents explore within their assigned cells with penalties for deviations \cite{solanas_coordinated_2004}, \cite{bramblettCoordinatedMultiAgentExploration2022}. An earlier alternative is clustering-based allocation, using K-Means to partition frontiers into groups by optimizing a similarity criterion, typically the intra-cluster variance. Cluster centroids are initialized either randomly \cite{bramblettCoordinatedMultiAgentExploration2022}, \cite{noauthor_new_2011}, \cite{goodwin_k-means_2022}, from agent positions \cite{lee_multi-agent_2025}, from map center \cite{solanas_coordinated_2004}, or via K-Means++ to select well-separated initial seeds that ensure more predictive outcomes after convergence. 

K-means variants, unlike Voronoi diagram, relax strict proximity to agents by forming cells based on inter-node relationships (e.g., proximity, connectivity). As a result, cells are not directly tied to UAVs and require a subsequent assignment stage based on utility criteria (cf. \cref{ssec_fbe}). Thus, the UAV-cell assignment can be addressed through greedy \cite{solanas_coordinated_2004}, \cite{bramblettCoordinatedMultiAgentExploration2022}, \cite{goodwin_k-means_2022} or Minimum Position ranking \cite{bautinMinPosNovelFrontier2012}, which prioritize robots per frontier rather than frontiers per robot. Global optimization methods, such as the Hungarian algorithm \cite{noauthor_new_2011} also minimize the overall assignment cost across all agents. K-means variants are more sensitive to topological changes, particularly in sparse graphs. In contrast, the Voronoi diagram provides a more stable solution by consistently allocating nearby frontiers to UAVs, guiding them in fixed directions with greater resistance to abrupt objective deviations, such as when another agent discovers a high-density area.

\subsection{Graph-based partitioning}

MRTA largely rely on euclidean distance assumptions, which poorly capture obstacle-constrained environments. Some works attempt to mitigate this limitation by incorporating obstacle-aware penalties \cite{tang_novel_2020}. Building on this idea, recent literature construct graphs \cite{zhangMRTopoMapMultiRobotExploration2022} where frontiers are nodes, carrying exploration state, and edges encode traversal costs derived from real-time shortest-path planners such as RRT, A* \cite{gaoMeetingMergingMissionMultirobotCoordinate2022}, Dijkstra \cite{dong_fast_2024}, \cite{dingBalancedCollaborativeExploration2025}. Communication volumes are reduced, replacing the exchange of voxel-level map representations by edge and nodes updates, sometimes enriched with semantic labels. 

The Generalized Voronoi Diagram (GVD), referred to as Voronoi hereafter, extends the classical Voronoi diagram using geodesic distance for a topology-driven partition \cite{dong_fast_2024, dingBalancedCollaborativeExploration2025}. Similarly, adaptations of K-means have been proposed to better handle graph-structured environments, using K-medoids \cite{lee_multi-agent_2025} or non-Euclidean distances and metrics adaptable to sparse frontier connectivity \cite{sieranoja_adapting_2022}. Unlike a centroid which is defined as the mean position of a set of nodes in Euclidean space (i.e., the average of their coordinates), a medoid is an actual node that minimizes total distance on the graph, making it suitable for non-Euclidean pairwise distances. Instead of relying solely on Euclidean distance, K-medoids leverage graph-theoretic measures such as PageRank centrality, which recursively quantifies node importance by recursive ranking, capturing global influence \cite{hajijPageRankKMeansClustering2021}.

Finally, few authors extend graphs beyond frontiers by incorporating agent positions, previously visited free-space locations \cite{gao_improved_2018}, \cite{dong_fast_2024}, and farther neighboring frontiers, to guide exploration toward map expansion \cite{dingBalancedCollaborativeExploration2025}, \cite{yang_active_2024}.

\subsection{Exploration under limited communication}

Most papers assume ideal communication conditions with continuous and reliable updates among agents, disregarding relative positions or adversarial factors like jamming or obstacles. Others maintain network connectivity constantly but sacrifice reactivity to link losses and limit swarm expansion for rapid mission completion. Only a quarter of the cited studies address robustness to uncertain communication and fully distributed MRE, leveraging scheduled rendezvous \cite{gao_improved_2018}, \cite{pongsirijindaMEFExploreCommunicationConstrainedMultiRobot2025} or opportunistic encounters to increase shared knowledge (e.g. map merging) and ensure more consistent offline decisions \cite{ribeiroEfficient3DExploration2024}, \cite{dingBalancedCollaborativeExploration2025}, \cite{yang_active_2024}. Decision-making may still remain centralized, when the whole swarm or connected subgroups operates, relying on auction mechanisms \cite{bramblettCoordinatedMultiAgentExploration2022} or leader election \cite{lupertoMultirobotRendezvousCommunicationrestricted2025}, \cite{gao_improved_2018}. We employ fully distributed strategies, where each robot decides using punctually shared, time-stamped observations and estimated exploration state.

\section{METHODS}
\label{sec_methods}

\subsection{Problem formulation}
\label{sec_pbstatement}

The environment, defined over a 3-D static fixed volume, is discretized into a grid of regularly-spaced cubic  subregions, referred to as Grid Exploration Subregions \subregionsName. Each \subregionsName ~is represented by its center, which defines its position, and undergoes the following stages~:

\begin{enumerate}
    \item Undiscovered
    \item\label{stage:frontier} Discovered, to be explored
    \item Discovered, fully explored or deemed infeasible
\end{enumerate}

We note that in stage~\ref{stage:frontier}, \subregionsName ~matches the state-of-the-art definition of frontier nodes introduced in \ref{ssec_fbe}.

We consider a team of $\numUAVs$ homogeneous Unmanned Aerial Vehicles (UAVs) launched near a common initial position. Each UAV is assumed to have access to accurate localization within the environment. During exploration, the UAVs maintain in parallel a consistent graph as a topological representation of the environment, encapsulating data on the exploration state, following \cite{dong_fast_2024}. For a UAV $\indexUAV$, the graph $\graphUAV = (\graphNodeSet, \graphEdgeSet)$ is composed of a node set $\graphNodeSet$ and an edge set $\graphEdgeSet$. The nodes $\node$ in $\graphNodeSet$ belong to three distinct types~:

\begin{itemize}
    \item $\vFTypeOne$ : Frontier nodes $\nodeFrontier$, representing discovered and explorable \subregionsName, i.e., in Stage~\ref{stage:frontier}.
    \item $\vHType$ : History nodes $\nodeHistorical$ which are sparsely distributed, free-space way-points previously reached by UAVs.
    \item $\typeUAV$ : UAV nodes $\nodeAgent$, representing each UAV
\end{itemize}

All nodes in the graph can be denoted as :
\[
\graphNodeSet = \{ {\node_{type}}_j \,|\, type \in Types = \{\vFTypeOne, \typeUAV, \vHType\}, j \in \mathbb{N} \}.
\]

For simplicity, we treat $\nodeAgent$ nodes as mobile $\nodeHistorical$. Accordingly, edges are defined between two $\nodeHistorical$ nodes ($\edgeHH$) or between a $\nodeHistorical$ node and a $\nodeFrontier$ node ($\edgeHF$), with each $\nodeFrontier$ connected to one or more $\nodeHistorical$ parents. Each $\nodeHistorical$ and $\nodeFrontier$ constructs a Dijkstra-based shortest-path tree over the free space to establish connectivity with other nodes. In our implementation, edges are explicitly built following the procedure detailed in \cite{dong_fast_2024}. This results in a dynamically maintained graph in which edges are created, removed, or updated according to changes in obstacle-aware connectivity.

UAVs exchange data opportunistically within a communication range $\commRadius$ (in meters) and relay the latest received system information. Therefore, messages encapsulate each UAV's current position and its $\nodeFrontier$ goal, each \subregionsName ~current stage and $\nodeHistorical$ locations. Note that nodes are shared among agents, whereas edge (weights and endpoints) are computed locally and not communicated, independently of communication range and protocol. We assume noise-free data sharing and seamless integration of maps and positions during merging.

\subsection{System Overview}

\begin{figure}[!t]
    \centering
    \includegraphics[width=\columnwidth]{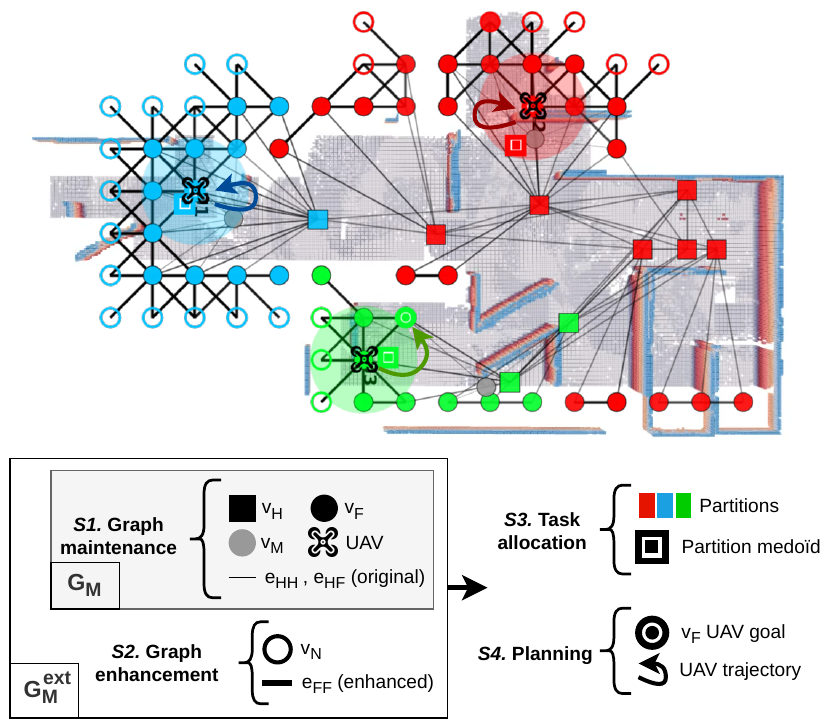}
    \caption{Overview of the proposed MRTA system, from an illustrative run}
    \label{fig:sys_overview}
\end{figure}

\par
Our method overview is illustrated in \cref{fig:sys_overview} and detailed in \cref{alg:async_framework}. At each step $S$, each UAV relies on the last known data on the system, swarm and mission progress, which may vary in freshness. The volumetric mapping module integrates the estimated pose and point cloud data to update the 3-D volumetric map, as shown in the background of the figure. The updated map is transmitted to the local ground station for visualization purposes only. Each UAV independently maintains its graph by adding sensed \subregionsName ~and $\nodeHistorical$ then running a first Dijkstra’s algorithm over the environment's cost map to connect nodes (see \cref{fig:sys_overview} $S1$).
\par
Extension and allocation are executed sequentially and repeated every second since graph topology is constantly changing. First, $\graphUAV$ is extended into ${\graphUAVext}$ by adding neighboring \subregionsName ~candidates (see \cref{fig:sys_overview} $S2$ and \cref{ssec_graph}). When using partition based on geodesic distances (see S3), we compute all-pairs shortest paths (APSP) using a second Dijkstra on the extended graph ${\graphUAVext}$. ${\graphUAVext}$ is then divided into $\numUAVs$ partitions using either Voronoï diagram or K-medoid adapted clustering (see \cref{fig:sys_overview} $S3$ and \cref{ssec_mrta}). UAV $\indexUAV$ only considers its assigned partition. Finally, the planning stage is decoupled from the allocation stage and triggered as \subregionsName ~are being explored. Each UAV $i$ selects a \subregionsName node to explore within its ${\graphUAVext}$ partition (see \cref{fig:sys_overview} $S4$ and \cref{ssec_plan}). $\graphUAV$ updates performed in step 1 and local UAV plan from step 4 are opportunistically shared with other agents while graph extension and partition are not communicated (cf. \cref{sec_pbstatement}).

\begin{algorithm}
\caption{Asynchronous UAV $i$ Exploration Framework}
\label{alg:async_framework}

\begin{algorithmic}[1]
\Require Graph $\graphUAV$, UAV node $\uavNode$, sensed \subregionsName ~S

\Function{UpdateGraph}{$\graphUAV$, $\uavNode$, $\mathcal{S}$}

    \If{$\forall \nodeHistorical,\ \|\uavNode - \nodeHistorical\| \ge X$}
        \State Add $v_H$ at position $\uavPose$
        \State Update edges with APSP (Dijkstra 1)
    \EndIf

    \ForAll{$s \in \mathcal{S}$}
        \State \textbf{if} $s$ reachable: connect to nearest $\nodeHistorical$; \textbf{else} discard
        \State \textbf{if} $s$ explored: remove from $\graphUAV$
    \EndFor

\EndFunction

\Function{TaskAllocation}{$\graphUAV$, $\uavPose$}
    \State Construct extended graph $\graphUAVext$ (see \cref{alg:ext_graph})
    \State Create distance matrix $\matrixAdj$ with APSP (Dijkstra 2)

    \State Divide $\graphUAVext$ into $\numUAVs$ partitions :
    \State \hspace{-2em} \textit{Method: closest to UAV (GVP) or medoid (K-medoids)}
    \State \hspace{-2em} \textit{Metric: node positions (Euclidian) or $\matrixAdj$ (Geodesic)}

\EndFunction

\While{free-space is not fully explored}

    \If{$\exists \nodeAgent^i ,\ \|\uavNode - \nodeAgent^i\| \leq \commRadius$}
        \State Exchange graph $\graphUAV$ with neighbors UAV $i$
        \State \Call{UpdateGraph}{$\graphUAV$, $\uavNode$, received \subregionsName ~$\mathcal{S}$}
    \EndIf

    \State \Call{UpdateGraph}{$\graphUAV$, $\uavNode$, sensed \subregionsName ~$\mathcal{S}$}
    \State \Call{TaskAllocation}{$\graphUAV$, $\uavNode$}
    \State Plan motion toward closest assigned node

\EndWhile

\end{algorithmic}
\end{algorithm}

\subsection{Hierarchical Graph-Frontier Planning}

\subsubsection{Task Allocation}
\label{ssec_mrta}

\par
\begin{figure}[!t]
    \centering
    
    \subfloat[Generalized Voronoï Diagram%
    \label{subfig_graphVor}]{
        \includegraphics[
            width=0.47\columnwidth
        ]{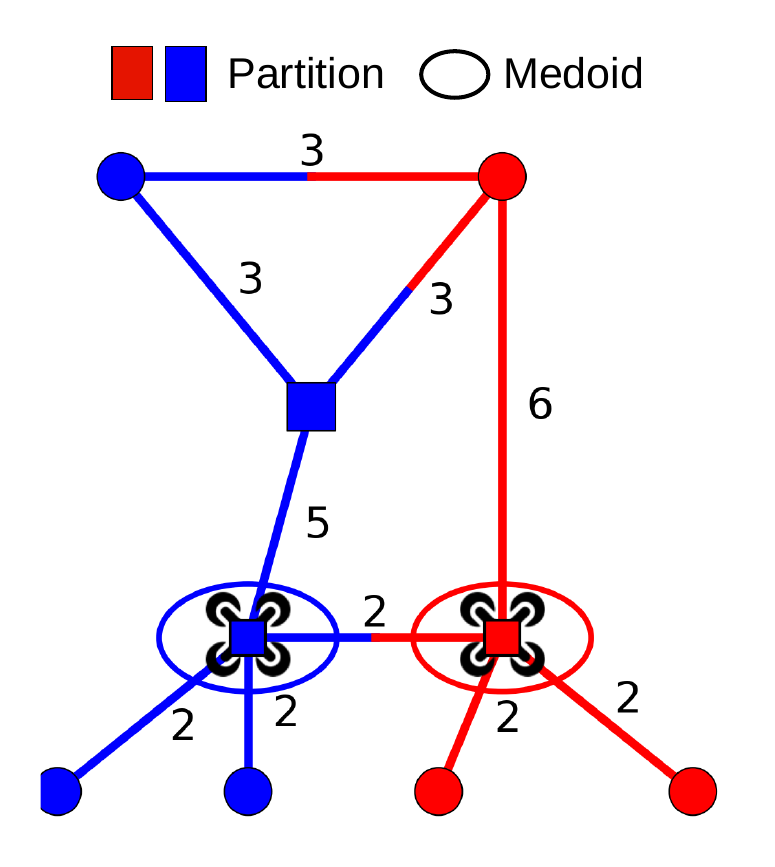}
    }
    \hfill
    \subfloat[K-medoid (closeness centrality)%
    \label{subfig_graphKmed}]{
        \includegraphics[
            width=0.47\columnwidth
        ]{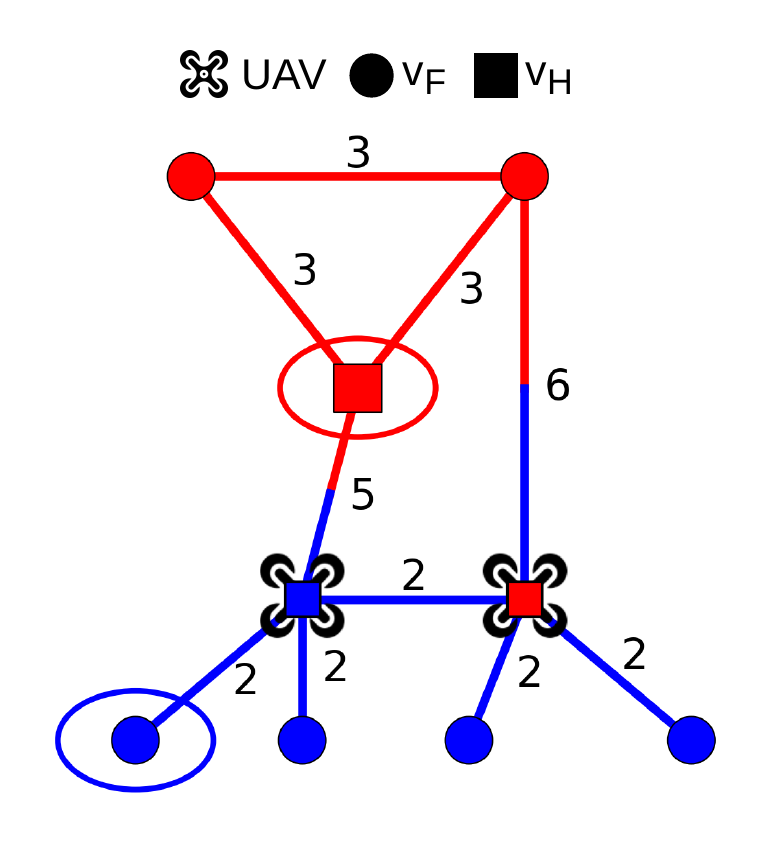}
    }
    
    \caption{Comparison of graph partitioning methods, with $\numUAVs=2$}
    \label{fig:part_methods}

\end{figure}

We compared our graph extension algorithm using two partitioning methods: Voronoi diagrams and the K-medoids variant of K-means. Detailed descriptions of both algorithms can be found in \cite{dong_fast_2024} for the Voronoi approach and in \cite{goodwin_k-means_2022} for K-medoids. As illustrated in \cref{fig:part_methods}, the main difference between the two methods lies in how the medoids are defined. K-medoids partitions the nodes into $\numUAVs$ clusters based solely on the graph structure, independently of the current UAV positions. In contrast, the Voronoi method directly uses the UAV positions as medoids by definition.

\par
We removed $\nodeAgent$ nodes from K-medoids clustering to limit local graphs divergence due to unreliable, rapidly outdated UAV position data. Unlike most state-of-the-art methods which use random medoids selection, we employ K-means++ strategy to initialize medoids to $\numUAVs$ farthest away nodes in $\graphUAV$. By selecting distant nodes initially, we aim to reduce initialization variance because these nodes are likely to be in different clusters and less prone to immediate changes in their connectivity within the graph compared to other nodes. We employ the closeness centrality method \cite{hajijPageRankKMeansClustering2021} for our analysis due to its focus on distance metrics. Unlike other centrality measures such as PageRank or between-ness centrality, closeness centrality assesses how easily a node can reach all other nodes in the graph, emphasizing shortest path relationships. 

Cluster are then attributed to UAVs. In the Voronoi approach, UAVs act as medoids, so assignment is inherently coupled with partitioning. In K-medoid, each medoid must be explicitly mapped to a UAV. A cost matrix of minimal distances between medoids and UAVs is computed, and the Hungarian algorithm performs optimal one-to-one assignment by minimizing total cost.

\subsubsection{Planning}
\label{ssec_plan}

\par
Once the nodes in $\graphUAVext$ are assigned to each agent, every UAV proceeds with sequential exploration of the \subregionsName ~corresponding to its allocated nodes. Each UAV explores the closest \subregionsName, where closeness is defined using a weighted distance that accounts for both the UAV position and its partition’s medoid \cite{solanas_coordinated_2004}. This approach avoids, on the one hand, planning toward overly peripheral regions of the partition, which would pull the UAV away from its cluster, and on the other hand, focusing only on centrally located regions, which would restrict the expansion of the explored space. 

\par
Newly discovered and connected $\nodeFrontier$ nodes are assigned by default to the discovering UAV if directly adjacent to its partition or improving relative position to its medoid. Otherwise, they remain unassigned and are redistributed during the task allocation phase,  preventing the UAV from being driven toward progressively distant regions outside its assigned partition. If exploration of a partition is completed before reallocation —either because the partition is empty or further obstacles detection makes the area inaccessible— the UAVs plan toward the \subregionsName ~still in $\graphUAVext$ attributed to other UAVs closest to its centroid.

\subsubsection{Complexity}
\label{sssec_complex}

\par
Johnson-Dijkstra’s algorithm is chosen for its lower complexity of $O((\numGraphEdge+\numGraphNode) \log(\numGraphNode)$, making it efficient for sparse graphs where the number of edges exceeds the number of nodes. However, we populate an adjacency matrix after applying Dijkstra’s algorithm to facilitate easy access to all pairwise distances, increasing the overall complexity to $O(\numGraphNode^2)$. This approach remains advantageous compared to algorithms like Floyd-Warshall, which have a higher complexity of $O(\numGraphNode^3)$. 

\par
The computational complexities for K-medoid and Voronoi are $O(\numGraphNode \numUAVs \kmeansSteps)$ and $O(\numGraphNode \, \log(\numGraphNode))$, respectively. However, since the adjacency matrix is computed in all cases for shortest paths, frequently accessed during planning, the overall system complexity remains quadratic. The primary advantage of Voronoi lies in its ability to perform individual assignments nearly in $O(\numGraphNode)$ (cf. \cref{ssec_partition_methods}). When applied to an adjacency matrix, Voronoi complexity becomes $O(\numGraphNode \numUAVs)$, placing it on par with K-medoid. Additionally, K-medoids performs clustering via distance minimization, resulting in an overall complexity of $O(\numGraphNode \numUAVs \kmeansSteps + \frac{\numGraphNode^2} {\numUAVs})$. Assuming $\kmeansSteps$ is constant, both algorithms exhibit near-linear scaling behavior for small swarms. However, when $\numUAVs$ increases, potentially approaching the number of nodes in the graph, the complexity degrades towards quadratic growth.

\subsection{Graph Extension}
\par
\begin{figure}[!t]
    \centering
    
    \subfloat[Conservation: disjoint $\nodeHistorical$ sets%
    \label{subfig_graphNo}]{
        \includegraphics[
            width=0.47\columnwidth
        ]{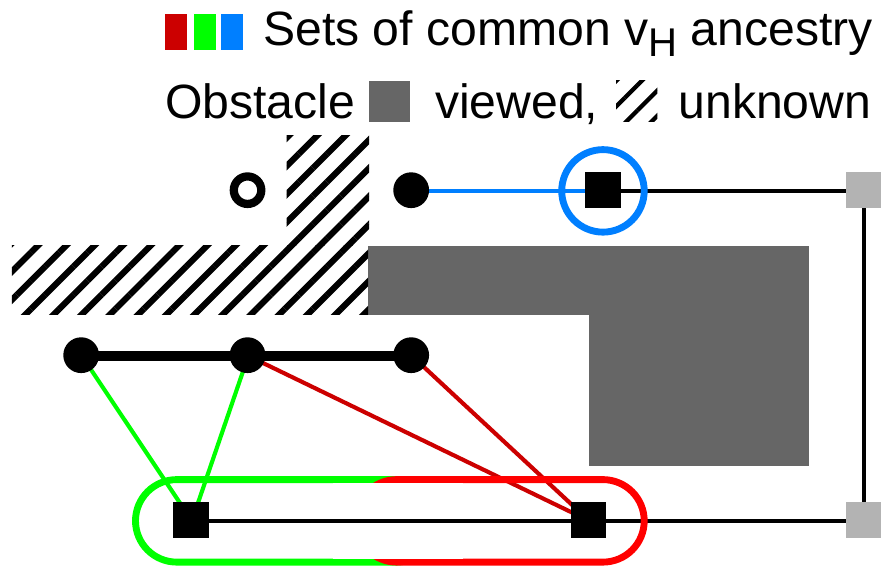}
    }
    \hfill
    \subfloat[Extension: intersecting $\nodeHistorical$ sets%
    \label{subfig_grapYes}]{
        \includegraphics[
            width=0.47\columnwidth
        ]{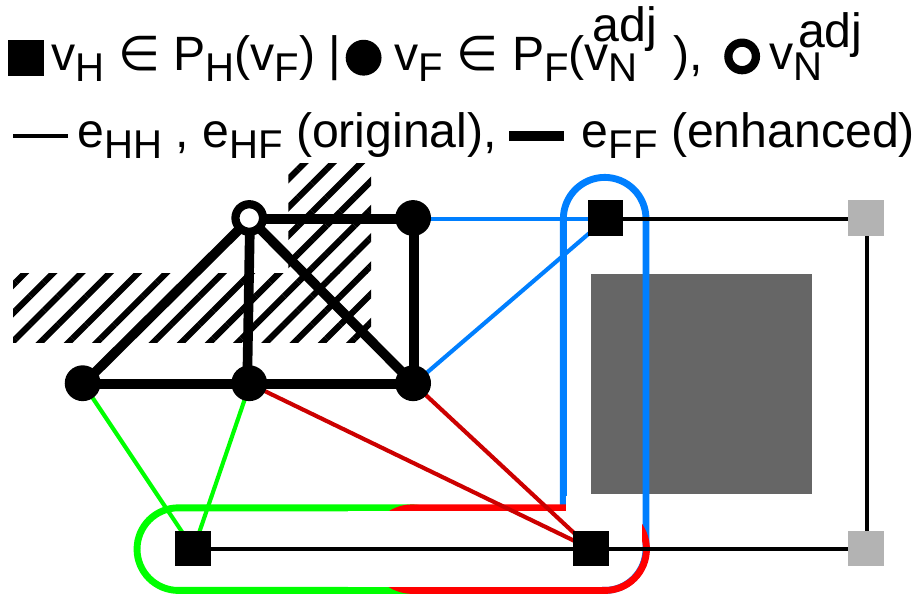}
    }
    
    \caption{Effect of $\nodeFrontierAdj$ parent-set connectivity on graph extension.}
    \label{fig:graph_extension}

\end{figure}

The proposed graph extension method temporary augment $\graphUAV$ by connecting selected \subregionsName ~during the distributed task partition process resulting in the extended graph $\graphUAVext$. Neighboring \subregionsName ~in stage 1, denoted $\nodeFrontierAdj$, are attached with associated edges to the overall structure to steer exploration toward space regions with higher \subregionsName ~concentration (see \cref{fig:sys_overview} step S2 and \cref{ssec_graph}). The intuition behind graph topological aware extension is highlighted in \cref{fig:graph_extension}. In \cref{subfig_graphNo}, edge addition is detrimental: one node (circled in blue) is topologically separated due to an obstacle and does not share common ancestry with the others. In this case, adding edges without topological consideration would introduce spurious shortcuts between otherwise distant regions of the graph. In \cref{subfig_grapYes}, in contrast, edge addition is beneficial, as the nodes share common ancestors and the added edges reinforce their existing structural proximity within the graph.
\par
\cref{alg:ext_graph} presents the proposed graph extension procedure. For each $\nodeFrontier$ in $\graphUAV$, we identify and establish parentage to its geographically adjacent \subregionsName ~in stage 1 neighbors, sharing a face or edge and not yet included in the graph and explored. We first gather all $\nodeFrontierAdj$, then, for each of them, identify the sets of $\nodeFrontier$ parent $\parentFset(\nodeFrontierAdj)$ with common $\nodeHistorical$ ancestry. As shown in \cref{subfig_grapYes}, if all parents belong to the same largest connected component (LCC), we add $\nodeFrontierAdj$ to the the graph and connect it to this parents via $\edgeFF$. Otherwise, $\nodeFrontierAdj$ is ignored as in \cref{subfig_graphNo}, to avoid creating spurious shortest paths. As mentioned above, no edges exist between $\nodeFrontier$ $\edgeFF$ in $\graphUAV$; They are introduced in $\graphUAVext$. The edge weight corresponds to the Euclidean distance between node positions, scaled by a penalty value $\penaltyFF>1.0$. This penalty accounts for uncertainty regarding obstacles since farther undiscovered nodes have unknown reachability from other nodes.

\label{ssec_graph}

\begin{algorithm}
\caption{Construct extended graph $\graphUAVext$}
\label{alg:ext_graph}

\begin{algorithmic}[1]
\State \textbf{Notation:} 
$\parentFset(\nodeFrontierZero)$: $\nodeFrontier$ parents of $\nodeFrontierZero$;
$\parentHset(\nodeFrontier)$: $\nodeHistorical$ parents of $\nodeFrontier$; 
$\mathcal{\parentHset}(\nodeFrontierZero)=\{\parentHset(\nodeFrontier)\mid \nodeFrontier\in \parentFset(\nodeFrontierZero)\}$.

\State $\graphUAVext \gets \graphUAV$

\ForAll{$\nodeFrontier \in \graphUAV$}
    \ForAll{$\nodeFrontierAdj \in$ \Call{GetAdjacent\subregionsName}{$\nodeFrontier$}}
        \State $\parentFset(\nodeFrontierAdj) \gets \nodeFrontier$,\quad $\mathcal{\parentHset}(\nodeFrontierAdj) \gets \parentHset(\nodeFrontier)$
    \EndFor
\EndFor

\ForAll{$\nodeFrontierAdj$}
    \State $\mathcal{U} \gets \parentFset(\nodeFrontierAdj) \cup \parentHset(\nodeFrontierAdj)$

    \If{$\Call{LargestConnectComponent}{\mathcal{U}}=\mathcal{U}$}
        \State $\graphUAVext$ nodes $\gets \nodeFrontierAdj$

        \ForAll{$\nodeFrontier \in \parentFset(\nodeFrontierAdj)$}
            \State $length_e \gets \penaltyFF \, \|\nodeFrontier - \nodeFrontierAdj\|$
            \State $\graphUAVext$ edges $\gets (\nodeFrontierAdj, \nodeFrontier, length_e)$
        \EndFor
    \EndIf
\EndFor

\end{algorithmic}
\end{algorithm}

\section{EXPERIMENTS AND RESULTS}
\label{sec_exp_result}

\subsection{Experimental setup}

\par
We evaluate the algorithms using total exploration time $T_{tot}$, average UAV traveled distance $D_{avg}$, and the standard deviation of total traveled distance across agents $D_{std}$. Times are reported in seconds, and distances in meters. These metrics capture operational constraints by promoting fast exploration while limiting unnecessary motion and imbalance between UAVs. We additionally introduce $\tau$, the average computation time of a graph partition during exploration. To evaluate performance under a fixed communication range and opportunistic communication, i.e., without connectivity strategy, we introduce four additional metrics. First, the connectivity ratio $\rho_{conn} = \frac{1}{T_{tot} k} \sum_{0}^{T_{tot}} n^{r}_{\text{conn}}(t)$ is the normalized average number of communication partners per UAV over a run $r$. Second, we define $B_{tot}$ as the total volume of data exchanged by a UAV during the run, measured in megabits. We then quantify both intra-UAV stability and inter-UAV consensus of partition medoid estimates under varying connectivity conditions.

\begin{figure}[h]
    \centering
    \makebox[\columnwidth]{%
        \subfloat[Maze, $2560 m^3$ \cite{dong_fast_2024}]{%
            \label{subfig_envMaze}
            \includegraphics[width=0.42\columnwidth]{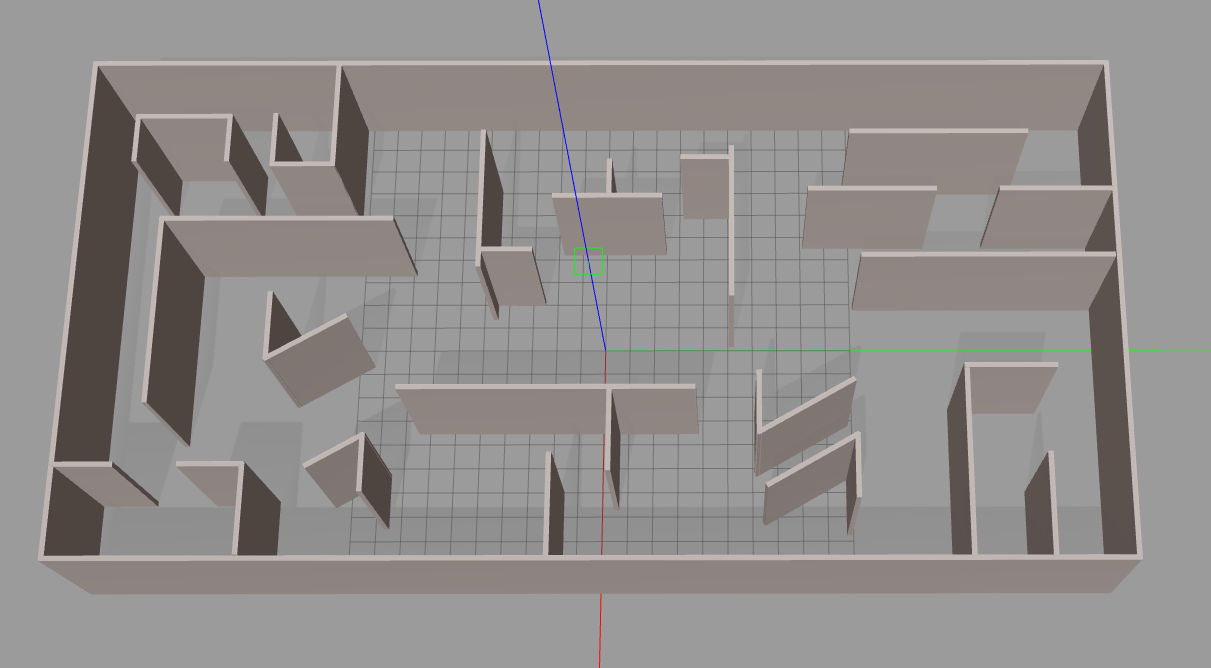}
        }
        \hfill
        \subfloat[Subterranean (SubT), $13710 m^3$]{%
            \label{subfig_envSubT}
            \includegraphics[width=0.48 \columnwidth]{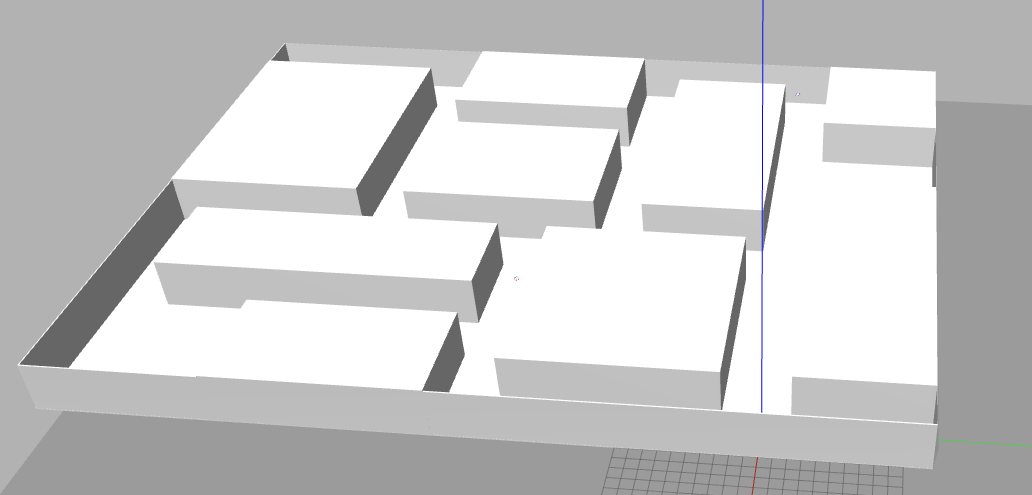}
        }
    }
    \caption{Simulation environments}
    \label{fig:simu_env}
\end{figure}

\par

We first evaluate the stability of each UAV’s assigned cluster medoid trajectory. This metric is only meaningful for K-medoids-based approaches, since in Voronoi partitioning the medoid trajectory coincides with the UAV trajectory. Let $\mathbf{x}^{r,i}(t_k) \in \mathbb{R}^3$ denote the medoid position of UAV $i$ at time $t_k$ in run $r$, and let $c^{r,i}(t_k) \in \{0,1\}$ indicate whether UAV $i$ is connected to at least one other agent over the time window $[t_k-\tau, t_k]$. Stability is defined as the trace of the covariance of the medoid trajectory over the sliding window:
\begin{equation}
V_{s}^{r,i}(t_k) = \mathrm{Tr}\left(\mathrm{Cov}\left(\{\mathbf{x}^{r,i}(t_j)\}_{t_j \in [t_k-\tau, t_k]}\right)\right).
\end{equation}

We additionally quantify inter-UAV consensus by measuring the dispersion of medoid estimates across agents. At each time $t_k$, every UAV computes a medoid position for a given UAV $i$, and these estimates are averaged across all UAVs to obtain $\bar{\mathbf{x}}^{r,i}(t_k)$. The consensus metric is then defined as the temporal covariance of this averaged estimate over the sliding window:

\begin{equation}
V_{c}^{r,i}(t_k)
= \mathrm{Tr}\left(\mathrm{Cov}\left(\{\bar{\mathbf{x}}^{r,i}(t_j)\}_{t_j \in [t_k-\tau, t_k]}\right)\right).
\end{equation}

For both stability and consensus metrics, the mean is computed over disconnection intervals to quantify partition divergence :
\begin{equation}
V_{\phi, disc}^{r,i}
= \operatorname{mean}\{V_{\phi}^{r,i}(t_k)\;|\; c^{r,i}(t_k)=0\}, \phi \in \{s,c\}.
\end{equation}

Simulations are conducted using the ROS and Gazebo framework with the configuration settings detailed in \cref{table_config}. We select a 3-drone setup due to its low cost and inherent asymmetry, a prime configuration highlighting exploration performance variations. Each configuration is run 20 times; metrics are averaged over runs and variability is assessed using box and whisker plots (minimum, maximum, median, quartiles $Q1$ and $Q3$), with the mean also reported.
The original MRE algorithm relies on a Voronoi-based decomposition combined with global communications, without any form of graph extension, and the evaluation is restricted to relatively homogeneous, office-like environments, without assessing performance under various topology. We conduct experiments both in the original setup provided in \cite{dong_fast_2024} (\cref{subfig_envMaze}) and in a more complex map featuring multiple corridors and dead ends, to approximate subterranean conditions (\cref{subfig_envSubT}). The environments are intentionally kept with limited vertical extent (quasi-planar) for visualization purposes, although the framework fully supports 3D scenarios.

\begin{table}[t]
\caption{Simulation parameters}
\centering
\begin{tabular}{l c l c}
\toprule
Parameter & Value & Parameter & Value \\
\midrule
Number of UAVs & 3 & Voxel size & $0.001\, m^3$ \\
Average UAV velocity & $1.1\, m.s^{-1}$ & \subregionsName ~size & $8\, m^3$ \\
Communication radius & $6\, m$ & Sensing radius & $4\, m$ \\
\bottomrule
\end{tabular}
\label{table_config}
\end{table}

Our approach on graph extension is applied to two clustering methods: K-medoids clustering on graphs \cite{goodwin_k-means_2022} and the communication-constrained version of the original graph-based Voronoi method \cite{dong_fast_2024}. For graph extension, we compare against \cite{dingBalancedCollaborativeExploration2025}, which samples points uniformly over the graph within a sensing-range window for a Voronoi-based allocation. We also compare graph partitioning based on topological distances (i.e., edge weights) and based on euclidean distances between node positions. The notation and encoding scheme used throughout this work are summarized in \cref{tab:benchmark_notation}. Each configuration follows the naming convention \textit{Method--Communication--Graph extension}. Methods enhanced with our approach are indicated in bold in the legends and tables, while the others are represented with hatched box-plots.

\begin{table}[t]
\caption{Compared method labeling}
\centering
\begin{tabular}{lll}
\toprule
Category & Notation & Description \\
\midrule
\multirow{2}{*}{Method}
& K-medoid \cite{goodwin_k-means_2022}  & K-medoid partition + Hungarian \\
& Voronoi \cite{dong_fast_2024}    & Voronoi-based allocation \\
\midrule
\multirow{2}{*}{Communication}
& Glob (global)    & full-range \\
& Loc (local)     & range-limited \\
\midrule
\multirow{3}{*}{Graph extension}
& Std    & None \\
& \textbf{Ext (ours)}    & Topological connectivity-based \\
& Samp \cite{dingBalancedCollaborativeExploration2025}    & Sampling-based \\
\bottomrule
\end{tabular}
\label{tab:benchmark_notation}
\end{table}

\subsection{Results analysis}

\begin{figure}[h] 
    \centering 

    \subfloat[Maze]{%
        \label{subfig_maze_results} 
        \includegraphics[width=0.9\columnwidth]{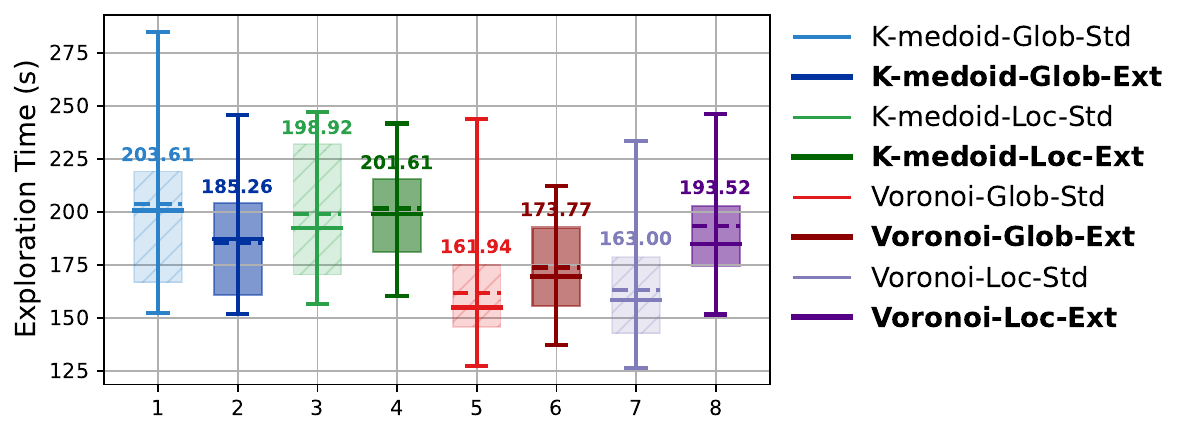} 
    } 

    \vspace{0.3cm}
    \subfloat[Subterranean (SubT)]{%
        \label{subfig_subt_results} 
        \includegraphics[width=0.9\columnwidth]{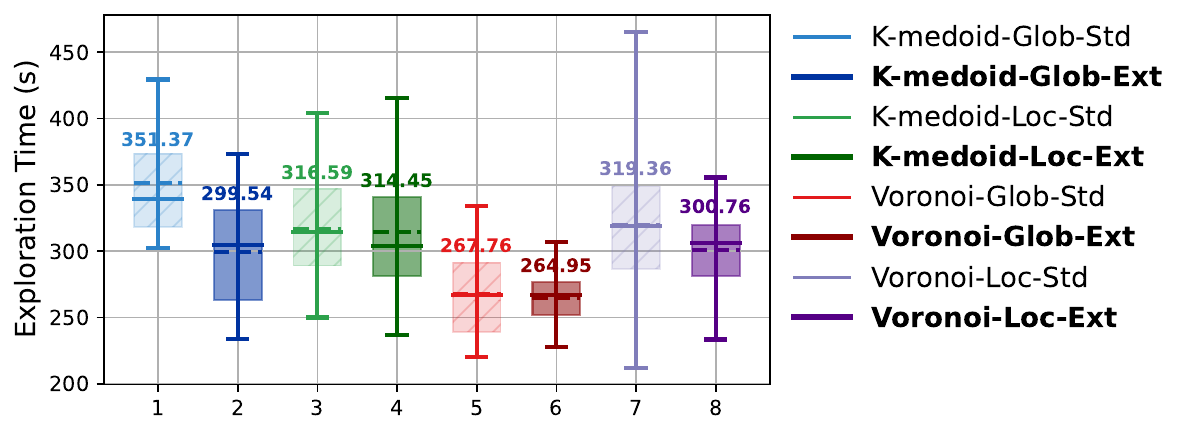} 
    } 

    \caption{Comparison of environments and communication scope on total exploration time required  to explore 95\% of the map.} 
    \label{fig:envs_boxes_fig} 
\end{figure}

\begin{table*}[t]
\caption{Performances comparison of methods under limited communication (Loc), in environment SubT}
\label{tab:results_scope}
\centering
\small
\begin{tabular}{cc||cc|cccccccc}
\toprule
\multirow{2}{*}{Scope} & \multirow{2}{*}{Method} &
\multicolumn{2}{c|}{Global} &
\multicolumn{8}{c}{Local} \\
\cmidrule(lr){3-4}\cmidrule(lr){5-12}

& & $\overline{T}_{tot}$ & $\overline{D}_{avg}$
& $\overline{T}_{tot}$ & $\overline{D}_{avg}$ & $\overline{D}_{std}$ & $\overline{\rho}_{conn}$ & $\overline{B}_{tot}$
& $\overline{V}_{s,\mathrm{disc}}$
& $\overline{V}_{c,\mathrm{disc}}$
& $\overline{\tau}$ \\
\midrule\midrule

\multirow{2}{*}{Std}
& Voronoi
& \underline{267.8} & \textbf{291.1}
& 319.4 & 352.7 & 9.8 & 0.14 & 252.3 & -- & -- & $3 \times 10^{-4}$ \\
& K-medoid
& 351.4 & 360.7
& 316.6 & 355.5 & 9.8 & 0.18 & \underline{221.8} & 59.1 & \underline{244.0} & 1.1 \\
\midrule

\multirow{2}{*}{Ext (ours)}
& Voronoi
& \textbf{265.0} & \underline{295.0}
& \underline{300.8} & \underline{335.9} & \textbf{4.9} & 0.17 & \textbf{215.3} & -- & -- & 1.1 \\
& K-medoid
& 299.5 & 316.0
& 314.5 & 364.3 & \underline{5.5} & 0.18 & 252.8 & 76.6 & \textbf{241.7} & 1.1 \\
\midrule

\multirow{2}{*}{Samp}
& Voronoi
& 307.1 & 316.3
& \textbf{298.6} & \textbf{334.7} & 6.6 & 0.21 & 225.3 & -- & -- & 1.1 \\
& K-medoid
& 273.6 & 305.2
& 319.7 & 357.0 & 18.5 & 0.17 & 229.0 & 56.9 & 259.6 & 1.1 \\
\bottomrule
\end{tabular}
\end{table*}

\subsubsection{Environment dependence}

\par
In \cref{subfig_maze_results}, for the maze environment, Voronoi and K-medoid exhibit similar performance under both local and global communication regimes, occasionally achieving slightly better results in the local setting, with or without graph extension. This suggests the exploration partitioning process is only weakly affected by the communication radius; even under limited communication, the swarm converges to similar exploration strategies with no significant performance loss. Such behavior highlights the robustness of Voronoi in sparse, unclustered environments, while suggesting that communication-limited settings provide limited additional insight in these conditions.

\par
In contrast to Maze, for the SubT environment in \cref{subfig_subt_results}, most partition methods perform better, as expected, under global communication, except K-medoids, which shows higher variability. This stems from the low number of frontiers, leading to sparse graph connectivity and strong bottlenecks due to insufficient node density at the boundaries. As a result, clusters shift rapidly across UAVs, causing frequent changes in partition structure.

\par
In \cref{tab:results_scope}, we observe similar performances for Voronoi-Loc with both extension methods, which is likely related to the topology of the environment. In environments with thin walls or tight separations, clustering nodes geographically close but topologically distant lead to inefficient backtracking. The SubT environment, however, is characterized by non-concave obstacles, wide corridors, and low-curvature turns. These properties create regions that are spatially extended while remaining clearly separated at their extremities, reducing ambiguity in exploration progression. 

\par
In SubT environments structural complexity including dead-ends and crossroads limits partitioning effectiveness making robust graph extension beneficial. Subsequent evaluations focus on the SubT environment.

\subsubsection{Effect of graph extension}
\par
Overall, graph extension improves both partitioning methods. In global settings, topological extension preserves Voronoi performance, while sampling-based extension improves K-medoids stability, reducing $\overline{T}{tot}$ and $\overline{D}{avg}$ by ~$15\%$ (see \cref{tab:results_scope}). In local settings, topological extension maintains K-medoids performance, while both extensions improve Voronoi, with gains of ~$20s$ in $\overline{T}{tot}$ and ~$20m$ in $\overline{D}{avg}$.

\par
Graph extension stabilizes K-medoid clusters shape, position, and assignment. Drones prioritize broader exploration due to continuously integrating information from other UAVs located in different regions of the environment. As environmental knowledge increases, the system becomes less driven toward exploration and increasingly focused on thoroughly exploring remaining $\node_{\vFTypeAny} \in \graphUAVext$ within known regions.

\subsubsection{Local dynamics}

\par
K-medoids, which we modified for stable graph partitioning, performs similarly to Voronoi under local and opportunistic communication, and both remain insufficient to match the performance of their global counterparts. In contrast, while K-medoids shows little benefit from graph extension, Voronoi is consistently improved by both extension methods, reducing $\overline{T}_{tot}$ by ~20 s and $\overline{D}_{avg}$ by ~18 m per UAV. We also observe that $\overline{D}_{std}$ is significantly lower with topological extension for both Voronoi and K-medoids, indicating that UAVs tend to travel more similar distances during exploration, independently of the frontiers exploration rate.

\par
Although $\overline{\rho}_{conn}$ is similar across configurations, it only reflects average connectivity, not contact timing or data volume per encounter. In practice, encounter frequency and spacing influence data exchange, as longer independent exploration increases accumulated information. Thus, despite similar $\overline{\rho}_{conn}$, the extended Voronoi approach reduces $\overline{B}_{tot}$ while maintaining better performance through more even spatial distribution and reduced travel to new areas. 

\par
K-medoids performs better when $\overline{V}_{s,\mathrm{disc}}$ is high, as connected communication uniformly propagates information across the map, leading to more centrally located medoids and increased back-and-forth UAV motion. The sampling-based extension encourages early spatial dispersion and improves the ratio of explored area to traveled distance in the initial phase. However, in later stages, reduced awareness of other agents’ states and explored regions limits efficient discovery of remaining areas, similar to standard graph configurations, with minimal redundancy.

\subsubsection{Allocation robustness}

\par
Across all categories and independently of the graph extension, K-medoid consistently yields higher total exploration times and higher average traveled distances than Voronoi. This confirms the greater robustness of Voronoi for swarm exploration in the considered settings.  The comparison of partitioning approaches has shown that the strength of one algorithm is the weakness of the other: K-medoids has a higher grasp on environment exploitation and UAVs coordination, while Voronoi pushes the swarm towards deeper exploration achieving higher performances. 

\par
We observed that, in the standard configuration without extension, Voronoi-based partitioning is only locally consistent in the short term during direct drone communication, but regions may evolve in parallel along similar directions in disconnected neighborhoods without synchronization. In contrast, K-medoid clustering can produce fragmented small clusters in previously unvisited areas, leading to biased exploration confined to residual regions or artificial assignment of drones to poorly representative zones. Both approaches rely on single-source information that lacks long-term validity.

\par
As mentioned before, under limited communications, no significant performance improvement is observed for K-medoid when applying the proposed extension in terms of exploration efficiency. However, our topological extension induces lower cluster-position divergence $\overline{V}_{c,\mathrm{disc}}$ than the sampling-based extension, while also achieving better consensus during disconnection than the standard baseline and indicating a more consistent cluster assignment over time.

\section{CONCLUSION}

This article presents a novel exploration graph extension method based on nodes parenting connectivity, to support distributed regions allocation during multi-agent exploration of unknown environments under limited-communication. Partitioning efficiency depends on environment topology and Euclidean distance can outperform topological metrics. In constrained environments, enriching the exploration graph under low communication rates is beneficial. It helps maintain deeper exploration, reduces overlap with other drones' inferred exploration zones, and mitigates performance degradation.

Future work includes increasing robustness to environmental topology and refining decision criteria, particularly regarding the trade-off between deep exploration and lateral area completion. A first direction to improve graph extension and local knowledge accuracy through parameter tuning, trajectory prediction and learning-based approaches. A complementary research direction is the development of recovery strategies, such as rendezvous, to increase the frequency and timing of information sharing within the swarm or toward an external operator, both for emergency intervention and improved shared knowledge supporting local decision-making.

 \addtolength{\textheight}{-12cm}   




\section{ACKNOWLEDGMENT}

We gratefully acknowledge Donovan Thaing for his help in performing experiments.

\bibliographystyle{IEEEtran}
\bibliography{main}

\end{document}